**Methodology for Comparing Machine Learning Algorithms for Survival Analysis**

Lucas Buk Cardoso[1,5,*], Simone Aldrey Angelo[2,*], Yasmin Pacheco Gil Bonilha[1], Fernando Maia[2], Adeylson Guimarães Ribeiro[3], Maria Paula Curado[4], Gisele Aparecida Fernandes[4], Vanderlei Cunha Parro[1], Flávio Almeida de Magalhães Cipparrone[5], Alexandre Dias Porto Chiavegatto Filho[2], Victor Wünsch Filho[2,3], Tatiana Natasha Toporcov[2]

*[1]Núcleo de Sistemas Eletrônicos Embarcados - Instituto Mauá de Tecnologia*
*[2]Epidemiology Department - Faculdade de Saúde Pública da Universidade de São Paulo*
*[3]Information and Epidemiology - Fundação Oncocentro de São Paulo*
*[4]Epidemiology and Statistics on Cancer Group - A.C. Camargo Cancer Center*
*[5]Electronic Systems Engineering Department - Escola Politécnica da Universidade de São Paulo*
*[*]These authors contributed equally*

## Abstract

This study presents a comparative methodological analysis of six Machine Learning models for Survival Analysis (MLSA). Using data from nearly 45,000 colorectal cancer patients in the Hospital-Based Cancer Registries of São Paulo, we evaluated Random Survival Forest (RSF), Gradient Boosting for Survival Analysis (GBSA), Survival SVM (SSVM), XGBoost-Cox (XGB-Cox), XGBoost-AFT (XGB-AFT), and LightGBM (LGBM), capable of predicting survival considering censored data. Hyperparameter optimization was performed with different samplers, and model performance was assessed using the Concordance Index (C-Index), C-Index IPCW, time-dependent AUC, and Integrated Brier Score (IBS). Survival curves produced by the models were compared with predictions from classification algorithms, and predictor interpretation was conducted using SHAP and permutation importance. XGB-AFT achieved the best performance (C-Index = 0.7618; IPCW = 0.7532), followed by GBSA and RSF. The results highlight the potential and applicability of MLSA to improve survival prediction and support decision making.

## Introduction

Colorectal cancer (CRC) represents one of the major contemporary oncologic challenges. In 2022, more than 1.9 million new cases and over 900,000 deaths were estimated worldwide, making CRC the third most diagnosed cancer and the second leading cause of cancer mortality (Bray et al., 2024). In Brazil, 45,630 new cases are estimated annually, making it the second most frequent tumor in both sexes, surpassed only by prostate and breast (INCA, 2022). The disease shows pronounced prognostic heterogeneity: five-year survival for early-stage disease exceeds 90%, whereas metastatic cases have survival below 20% (American Cancer Society, 2025). Given the marked rise in incidence and mortality in recent decades, the application of analytical methods able to assess and predict survival for these patients has become increasingly important.

Survival analysis techniques are fundamental statistical tools for studying time-to-event outcomes: they accommodate censoring — for example, loss to follow-up — and allow estimation of survival probabilities over a defined period. Among the most widely used approaches is the Cox proportional hazards model (Cox, 1972), recognized for its robustness and interpretability, although it relies on the proportional hazards assumption, which is not always observed in real-world clinical data (Kleinbaum & Klein, 1996). When these assumptions are violated, alternative methodologies that can handle greater complexity and heterogeneity become necessary (Bustamante-Teixeira, Faerstein & Latorre, 2002).

Machine learning (ML) has emerged as a valuable alternative to overcome the limitations of traditional statistical approaches, with increasing applications in healthcare. Classification techniques such as Random Forest and XGBoost have shown strong performance in oncology, predicting early diagnosis, tumor staging, and risk stratification, often outperforming classical approaches by capturing the complex patterns inherent in clinical data (Li et al., 2025; Finn et al., 2023; Zhen et al., 2024). However, when applied to survival analysis, these models do not adequately account for censoring, which results in the exclusion of patients and a reduced sample size, thereby introducing selection bias and compromising survival rate estimation (Schober & Vetter, 2018). In a previous study, focused on predicting survival in colorectal cancer patients (Cardoso et al., 2023), XGBoost achieved strong performance; nevertheless, the exclusion of patients ranged from 7% to 27% for survival classification at one, three, and five years after diagnosis, due to censoring.

Given these considerations, the use of Machine Learning for Survival Analysis (MLSA) — algorithms specifically designed or adapted for survival data — becomes particularly relevant. This class of methods remains underexplored in oncological literature, especially regarding the application of time-dependent models for survival prediction in cancer (Cygu et al., 2023). Unlike traditional supervised models, these approaches incorporate censoring natively, preserving information from all patients in the cohort and avoiding the bias inherent in excluding censored cases (Huang et al., 2023). Although it is not a mandatory characteristic of MLSA models, there is evidence that ML techniques in general perform more effectively on large datasets, since the complexity of their algorithms requires substantial sample sizes to ensure stable estimates (Riley et al., 2020; Rajula et al., 2020). In this context, the Hospital Based Cancer Registries of the State of São Paulo (RHC/SP), maintained by the Fundação Oncocentro de São Paulo (FOSP), with its large extent and quality, provides an especially suitable setting for applying these methods.

Accordingly, this study proposes the application of six MLSA models — Random Survival Forest (RSF), Gradient Boosting for Survival Analysis (GBSA), Survival SVM (SSVM), XGBoost Cox (XGB-Cox), XGBoost AFT (XGB-AFT), and LightGBM (LGBM) — in a retrospective cohort of colorectal cancer patients using data from the RHC/SP (FOSP, 2025). The objective is to compare machine learning algorithms for survival analysis, examining methodological aspects such as evaluation metrics, hyperparameter optimization, and model interpretability, grounded in an extensive literature review, and to discuss their strengths and limitations when applied to clinical data. The practical application using the RHC/SP data also enabled a comparison of these models' performance with classification algorithms previously developed by our research group, as well as an assessment of their consistency and potential advances in survival prediction.

## Methods

*Study Population*

The RHC/SP, managed by FOSP, compiles sociodemographic and clinical information on cancer patients diagnosed since 2000, contributed by 81 public and private institutions across the state of São Paulo (FOSP, 2025). As of September 2024, the database contains more than 1.23 million analytic cases — patients without prior treatment. For this study, only colorectal cancer cases were selected.

It should be emphasized that the variables selected as model inputs relate to demographic data, service characteristics, and time from diagnosis to beginning of treatment; features concerning procedures performed or tumor recurrence were excluded. Therefore, patient information is available only until treatment initiation.

*Data Preparation*

The preparation of the dataset began with the selection of colorectal cancer cases — 90,625 patients with ICD codes C18, C19 and C20 — with the processing steps shown in Figure 1. Patients younger than 20 years, non-residents of the state of São Paulo, with undefined clinical staging or *in situ*, without anatomopathological confirmation, or patients who had undergone bone marrow transplantation were excluded. In addition, cases with morphologies other than adenocarcinoma (8140/3, ICD-O, 3rd ed.) were removed.

The necessary and specific adjustments to the dataset were made to ensure its consistency, optimize the application of predictive models, and refine the survival analysis. These adjustments involved, among other steps, calculating the interval in days between the first consultation and treatment, and between diagnosis and the start of treatment, categorizing them as ≤ 60 days, 61–90 days, > 90 days, and untreated. Finally, output variables were created: time (in months) from diagnosis to the date of last available information, and death from any cause (0 = No, 1 = Yes).

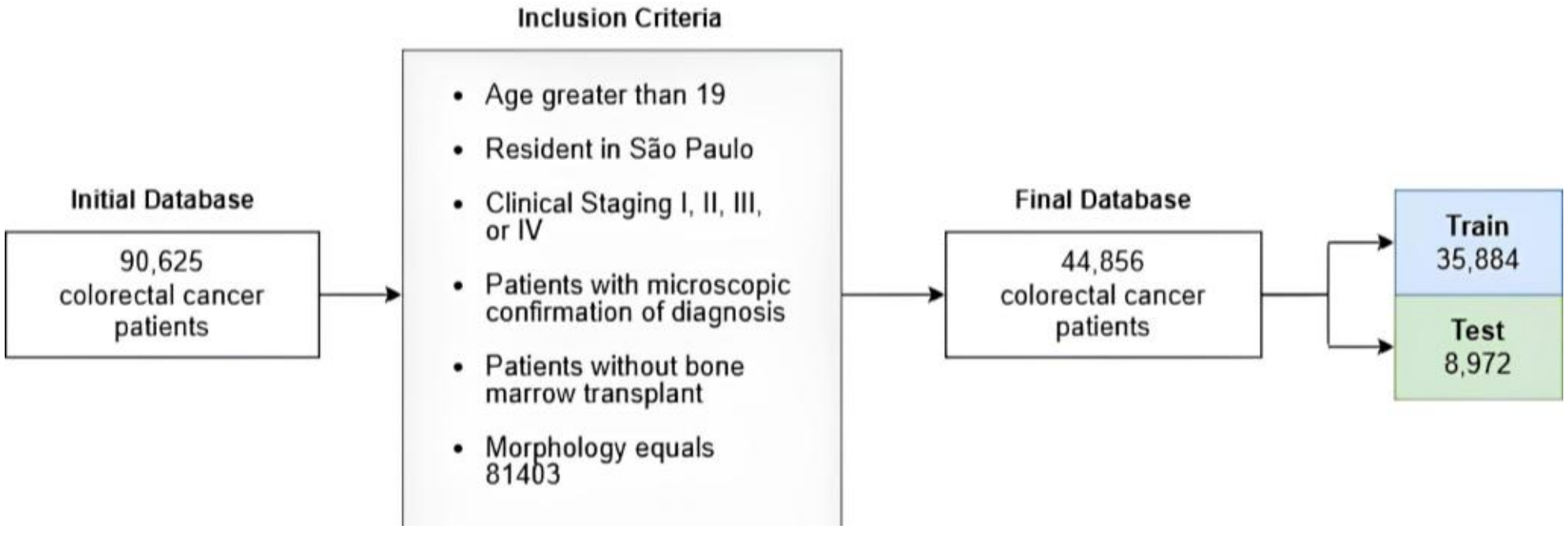

**Figure 1.** Diagram of the database processing steps. Description of selection criteria applied to the dataset with the final number of patients after adjustments, as well as the sizes of the datasets used for model training and testing.

The variables used as inputs to the ML models, along with their descriptions, are presented in Table 1.

**Table 1.** Description of the columns used as models' inputs.

| Feature | Description |
|---|---|
| INSTITU | Hospital code |
| ESCOLARI | Code for patient's education level |
| IDADE | Patient age |
| SEXO | Patient gender |
| IBGE | City code of patient's residence according to IBGE with check digit |
| CATEATEND | Category of care at diagnosis |
| DIAGPREV | Previous diagnosis and treatment |
| TOPO | Topography |
| EC | Clinical staging |
| ANODIAG | Year of diagnosis |
| DRS | Regional department of health |
| IBGEATEN | IBGE code of the healthcare institution where the patient was treated |
| HABILIT2 | Hospital Qualifications - Categories |
| DRS_INST | Hospital's Regional department of health |
| TRATCONS_CAT | Categorized difference in days between consultation and treatment dates |
| DIAGTRAT_CAT | Categorized difference in days between treatment and diagnosis dates |

*Classification Models*

Following the same approach used in a previous study by the group (Cardoso et al., 2023), which evaluated different classification models and found XGBoost to achieve the best performance for survival prediction, this model was retrained using the current dataset. That way, it became possible to compare the percentage of patients who survived according to the predictions of the traditional models — for 1, 3, and 5 years — with the curves produced by the survival analysis models, the focus of this work.

*Survival Models*

The six algorithms examined in this study employ distinct strategies to estimate time-to-event and were selected for their ability to handle censored data. Survival-specific models — RSF, GBSA, and SSVM — require the output variable to consist of two event-related elements: a status indicator (0 = censored, 1 = death) and a time component (time to death or last follow-up, for censored cases). For Machine Learning algorithms that lack native survival adaptations — XGBoost implemented under Cox or AFT parameterizations, and LGBM — modeling was conducted using only time-to-event as the target column. The status variable was incorporated as a weighting factor during training, assigning greater weight to deaths, as they provide more relevant and reliable information.

The dataset was split into training and validation sets: 80% of patients were used to train the algorithms and the remaining 20% to validate the models. The exact size for the training and test sets is shown in Figure 1.

Categorical variables (clinical staging and topography) were encoded ordinally to preserve potential hierarchical relationships between categories (Géron, 2022). Numeric variables were standardized using the z-score method, which rescales values to a mean of zero and a standard deviation of one, ensuring comparability across different ranges (Gareth et al., 2023). Since the Survival SVM model is sensitive to variable scaling (Van Belle et al., 2011), standardization was applied to all algorithms, even though tree-based models do not strictly require it since they split on thresholds (Hastie et al., 2005). This technique also improves numerical stability, accelerates convergence of optimization algorithms, and prevents variables with large magnitudes from dominating gradient and distance computations.

Among the algorithms evaluated in this study, the RSF is a decision-tree–based model applied to survival analysis (Ishwaran et al., 2008). Derived from Breiman's Random Forest (2001), RSF uses bagging and incorporates nonparametric estimators — such as the Nelson–Aalen — to infer the survival function and deal with censoring. Although robust, flexible and relatively interpretable, recent studies suggest that its predictive performance is inferior to boosting-based approaches (Ma et al., 2022; Wang et al., 2023).

GBSA is an extension of Gradient Boosting (Friedman, 2001) adapted for survival data that sequentially builds decision trees to accommodate censoring. Several works contributed for its development, including (Ridgeway, 1999; Benner, 2002; Hothorn et al., 2006; Schmid & Hothorn, 2008; and Chen et al., 2013). This study adopts Ridgeway's classical formulation (1999), which integrates the Cox partial likelihood as the loss function and iteratively fits trees to minimize that loss, thereby improving risk prediction, assigning higher weight on observations with lower likelihood. GBSA offers flexibility and accuracy for censored data but faces challenges such as substantial computational cost and an increased risk of overfitting.

The SSVM extends the Support Vector Machine (Cortes & Vapnik, 1995) to survival analysis; it was formalized by Van Belle et al. (2011) with contributions from Shivaswamy et al. (2007) and Van Belle & Lisboa (2014). SSVM is efficient in high-dimensional contexts and differs from the traditional SVM by incorporating censoring and using the concordance index (C-Index) as its performance metric. It adapts the margin-maximization principle through specialized loss functions. Nevertheless, SSVM often shows lower predictive performance than other MLSA models (Huang et al., 2023; Ma et al., 2022).

XGB-Cox and XGB-AFT are adaptations of XGBoost (Chen & Guestrin, 2016) for survival analysis. Based on Gradient Boosting, XGBoost builds decision trees sequentially and adds optimizations such as parallelization, second derivative of the loss function, and regularization, yielding high performance, robustness, and reduced training time. For survival modelling, the algorithm can be adapted with different loss functions — for example, the Cox partial likelihood to estimate hazard/risk functions, or an AFT-based loss (Barnwal, Cho & Hocking, 2022) that directly models survival time by assuming a probability distribution for the error term. There are other extensions, such as XGBLC (Ma et al., 2022) and EXSA (Liu et al., 2020).

LGBM (Ke et al., 2017) is a Gradient Boosting Decision Tree implementation designed for efficiency and scalability on large datasets. It employs techniques such as Gradient-based One-Side Sampling (GOSS), which prioritizes instances with larger gradients, and Exclusive Feature Bundling (EFB), grouping mutually exclusive features to reduce dimensionality. As LightGBM lacks a survival adaptation, the regression algorithm was used for the present study.

*Hyperparameter Optimization*

Hyperparameter selection was conducted with the goal of optimizing the performance of ML models. For this step, Optuna (Akiba et al., 2019) was used, in which three samplers — RandomSampler, TPESampler, and CmaEsSampler — were tested individually. After this comparison, for each model, the sampler that presented the best performance was selected, and the respective set of hyperparameters was adopted as the final configuration.

The RandomSampler employs a random search strategy in which hyperparameters are sampled independently from predefined distributions. Although this approach does not leverage information from previous trials, it is effective at exploring nonobvious regions of the search space and can help avoid local minima.

TPESampler is a Bayesian process-based method that optimizes hyperparameter selection using probabilistic models (Bergstra et al., 2011). At each iteration, the algorithm identifies the trials into "good" and "bad" groups, modeling the hyperparameter distributions for each group using kernel density estimators. New samples are generated preferentially from regions associated with better performance — thereby balancing exploration of new combinations with exploitation of the most promising areas.

Finally, the CmaEsSampler implements an evolutionary strategy that iteratively adapts a multivariate Gaussian distribution over the hyperparameter space (Hansen, 2016). In each generation, the sampler produces a population of candidates, evaluates their performance, and updates the mean and distribution covariance matrix to favor directions with higher yield (in this study, higher C-Index). This dynamic adaptation enables rapid convergence toward optimal regions, even in spaces with parameter correlations or stochastic noise.

The hyperparameter optimization phase employed those three sampling strategies, with each sampler evaluating 150 distinct parameter combinations and 10-fold cross-validation for every parameter set. Although these techniques increased computational time, they provided greater robustness in hyperparameter tuning.

*Models Validation*

Four metrics were used to evaluate the survival models: Concordance Index (C-Index), Inverse Probability of Censoring Weighting Concordance Index (C-Index IPCW), mean time-dependent Area Under the ROC Curve (mean time-dependent AUC), and Integrated Brier Score (IBS). The C-Index is a widely used survival metric that measures a model's discriminative ability — i.e., its capacity to correctly order individuals by survival time (Harrell et al., 1982). Values close to 1 indicate perfect concordance, whereas a value of 0.5 corresponds to random performance.

The C-Index IPCW is a more robust variant that corrects biases introduced by uninformative or covariate-dependent censoring. This adjustment is achieved via inverse-probability-of-censoring weights (IPCW), enabling a more accurate assessment in scenarios with censoring (Robins et al., 1994).

The area under the time-dependent ROC curve extends the traditional AUC — commonly used to quantify discrimination of binary classifiers (Hanley & McNeil, 1982) — to survival data. Introduced by Heagerty et al. (2000), this approach constructs a ROC curve for each follow-up time "t" and evaluates the model's ability to distinguish individuals who experienced the event by "t" from those who have not. Model performance is often summarized by the average of the initial and final estimation (values close to 1 indicate perfect discrimination, while 0.5 denotes random prediction).

The IBS is an extension of the Brier Score that assesses the accuracy of survival predictions over a time interval. Proposed by Graf et al. (1999), it integrates the squared prediction error over that interval while appropriately accounting for censored observations up to the time of censoring. Lower IBS values indicate better performance, with 0 representing a perfect prediction.

Importantly, the IBS is computed from the estimated survival function, and only RSF and GBSA provide native survival curves in their standard implementations — which yields greater fidelity in the temporal representation of risk. For SSVM, an approximate survival function was obtained by calibrating a Cox model, which allowed the generation of curves compatible with the theoretical assumptions and their respective IBS. In contrast, for XGB-AFT, XGB-Cox, and LGBM, despite adaptation attempts based on their theoretical formulations, it was not possible to achieve survival curves with sufficient fidelity to the data. As the lack of a reliable survival function approximation prevents consistent IBS computation, neither survival curves nor IBS values are presented for those algorithms.

Since the calculation of time-dependent AUC comes from risk predictions for patients, this metric was only used for the RSF, GBSA, SSVM, and XGB-Cox models — which directly predict risks, without the need to transform time estimates.

*Feature Importances*

SHAP (SHapley Additive exPlanations)) and Permutation Importance (PI) were used to assess the impact of predictive variables on the survival models — recognized reliable methods, particularly for tree-based ensemble models (Lundberg et al., 2017).

SHAP (Lundberg et al., 2017) combines game theory principles with local explanation techniques, assigning each feature an importance to individual predictions by computing local Shapley values (Shapley, 1953). The estimation considers all possible combinations of features and enables global interpretation by averaging individual contributions, thereby showing how each feature's values influence models' predictions.

Permutation Importance (Breiman, 2001) is a technique that quantifies the contribution of each feature to the performance of a model fitted to a tabular data set. The method randomly permutes the values of a single input column — leaving the others unchanged — and measures the degradation in a chosen performance metric, which reflects the model's reliance on that variable. Unlike SHAP, which explains individual predictions, PI provides only a global, model-level view of feature relevance.

*Ethical Considerations*

In accordance with CNS Resolution No. 510/2016, because this study used publicly available secondary data and did not contain patients' personal data, ethical approval from the Research Ethics Committee was not required.

**Results**

From the comparative validation of the MLSA models, XGB-AFT showed superior performance on the C-Index (0.7618), and the C-Index IPCW (0.7532), followed by the GBSA and RSF algorithms. GBSA achieved the best IBS (0.1552), and RSF presented the highest mean time-dependent AUC (0.8197). A summary of the metrics after hyperparameter tuning is presented in Table 2.

**Table 2.** Summary of results obtained by the best models after hyperparameter tuning.

| Models | *C-Index* | *C-Index IPCW* | *IBS* | *Mean AUC* |
|---|---|---|---|---|
| RSF | 0,7547 | 0,7489 | 0,1565 | 0,8197 |
| GBSA | 0,7588 | 0,7526 | 0,1552 | 0,8037 |
| SSVM | 0,7088 | 0,7036 | 0,1774 | 0,7458 |
| XGB - Cox | 0,7240 | 0,7184 | - | 0,7938 |
| XGB - AFT | 0,7618 | 0,7532 | - | - |
| LGBM | 0,7275 | 0,7211 | - | - |

Figure 2 shows the mean survival curves estimated by each model alongside the Kaplan-Meier curve, all computed on the validation set. Predictions of the classification models for 1, 3, and 5 years are indicated by "x" in the graph, it is important to highlight that the point estimations were lower than those from Kaplan-Meier: 0.6244 vs. 0.7623 (1 year), 0.4975 vs. 0.5410 (3 years), and 0.4223 vs. 0.4408 (5 years).

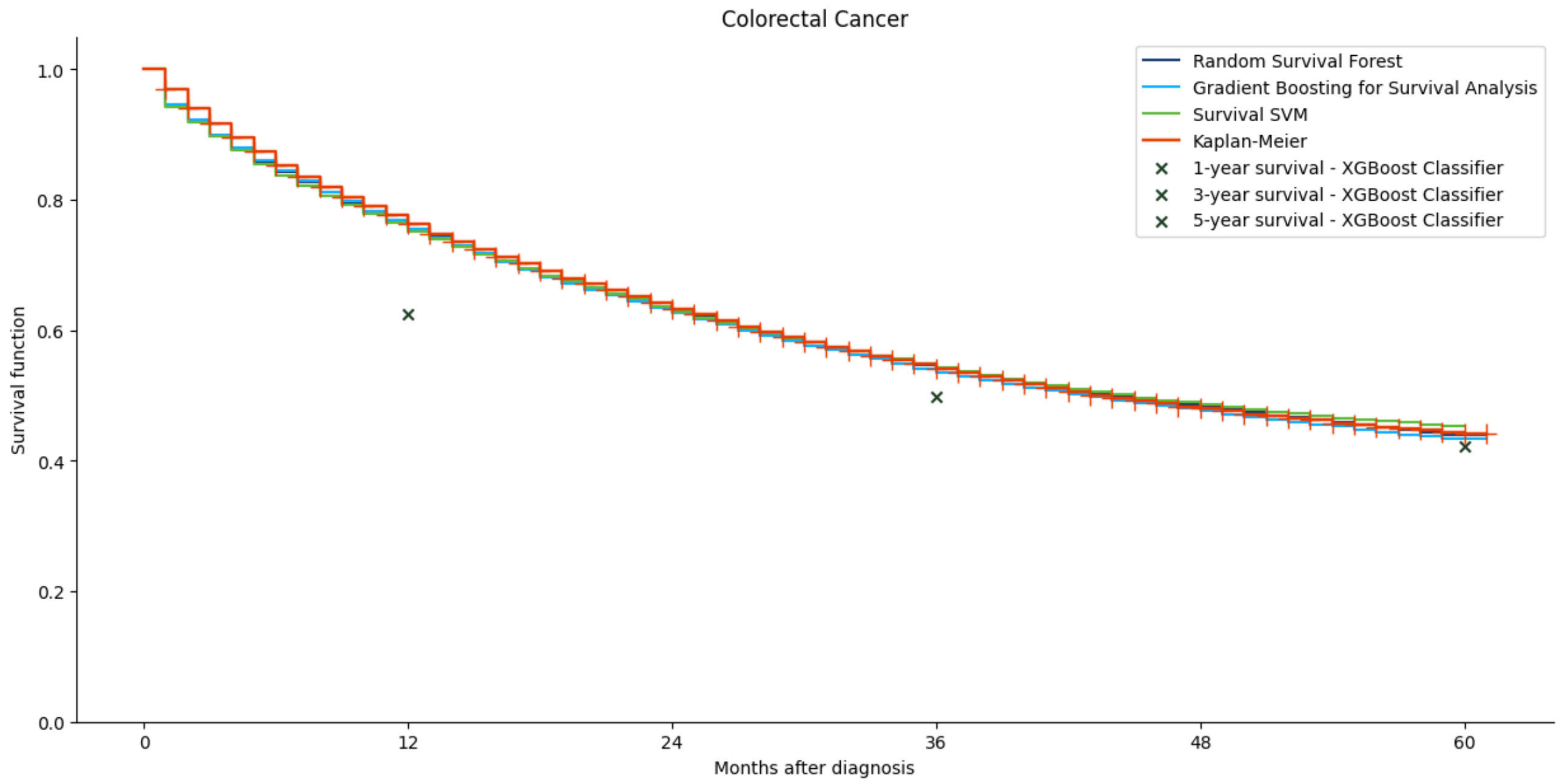

**Figure 2.** Predicted survival function curves. Survival predictions obtained by the classification models at 1, 3, and 5 years, in addition with the survival curves estimated by the MLSA models (RSF, GBSA, SSVM) and the nonparametric Kaplan–Meier estimator, plotted as a function of months since diagnosis.

The comparative efficiency analysis revealed substantial differences among the evaluated models — both in terms of performance and computational cost — corroborating the theoretical characteristics of each one. The algorithms XGB-AFT, GBSA, and RSF stood out for their high predictive efficiency compared with XGB-Cox, SSVM, and LGBM. When associating the computational cost with the hyperparameter optimization, GBSA exhibited substantially longer processing time than the other models, which showed moderate or even low resource demands. The relationship between each model's performance and its computational cost is presented in Table 3.

**Table 3.** Summary of the relationship between performance and computational cost for each model.

| Models | Performance | Computational Cost |
|---|---|---|
| RSF (Ishwaran et al., 2008) | *High* | *Average* |
| GBSA (Ridgeway, 2001) | *High* | *Most High* |
| SSVM (Van Belle et al., 2011) | *Low* | *Average* |
| XGB-Cox (Chen & Guestrin, 2016) | *Low* | *Low* |
| XGB-AFT (Barnwal, Cho & Hocking, 2022) | *High* | *Low* |
| LGBM (Ke et al., 2017) | *Low* | *Low* |

The most important features for XGB-AFT model according to each approach are shown in Figure 3. SHAP identified clinical staging (EC) as the top predictive feature, followed by age, hospital care category (CATEATEND), and interval between first consultation and treatment (TRATCONS_CAT). PI showed EC as the most important input by a wide margin; followed by TRATCONS_CAT and age.

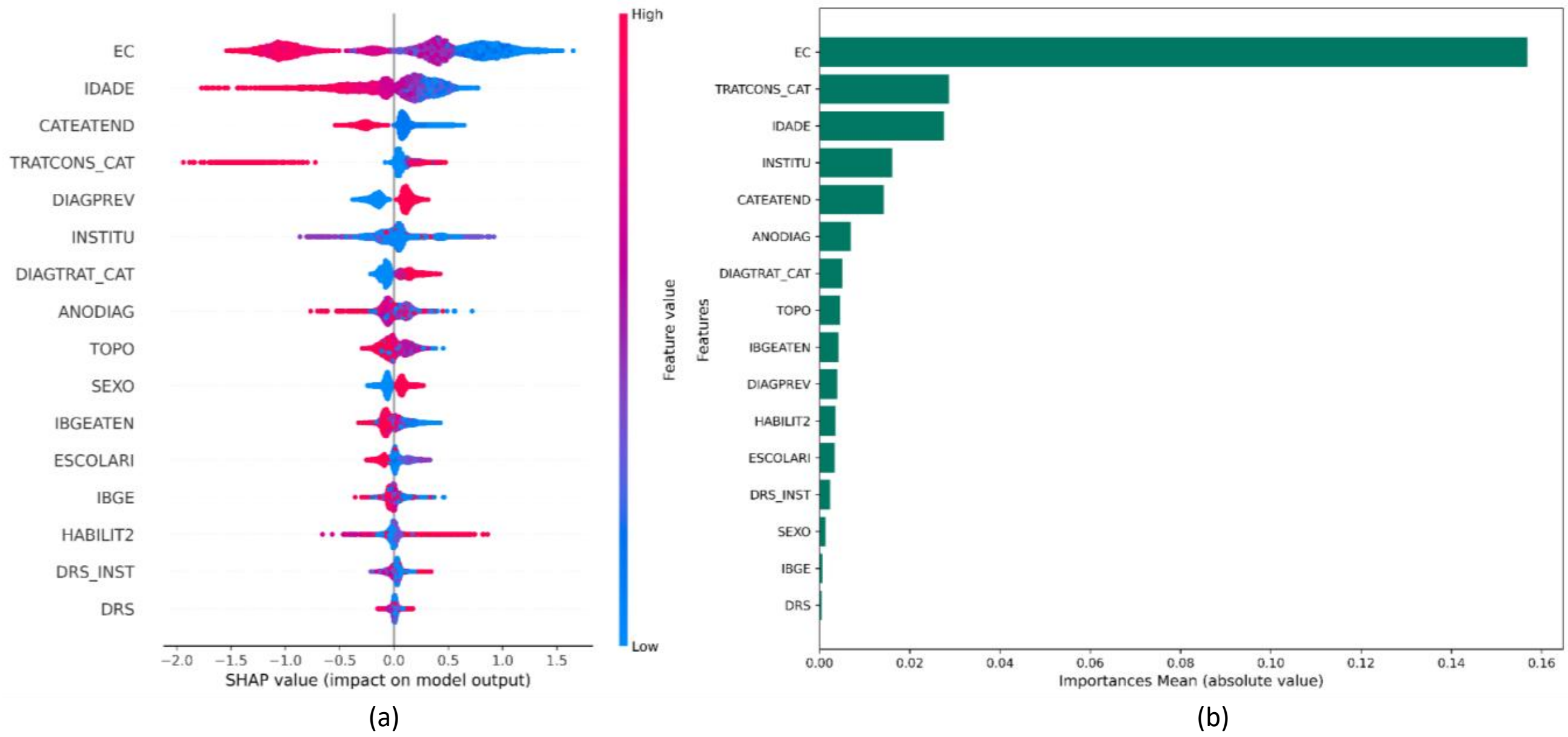

**Figure 3.** Feature importance analysis for XGB-AFT model. (A) SHAP summary plot showing the impact of features on the model's prediction. (B) Bar plot of mean absolute feature importance estimated by Permutation Importance (PI).

**Discussion**

This methodological study applied six ML models for survival analysis to a CRC cohort, enabling a detailed and thorough analysis of the main algorithms using a large, diverse, and complex database. The processes of variable encoding and standardization, use of specific metrics for survival analysis, hyperparameter optimization, and application of techniques for assessing variable importance in survival prediction were systematically addressed.

C-Index values ranged from 0.70 to 0.76, demonstrating the aptness of MLSA models to estimate survival probability. These algorithms excel at appropriately handling censored data, modeling the temporal dynamics of events, and estimating survival curves. Although more frequently discussed in international literature, their application remains incipient. The work by Wang et al. (2019) represents one of the first and most comprehensive reviews on MLSA application, highlighting the fragmentation of research across different disciplines and the scarcity of consolidated studies. The authors emphasize that, despite its potential, MLSA was, and in some aspects still is, in an early stage of exploration and systematization. Complementarily, Huang et al. (2023), in a scoping review, note that whereas there is growing interest in using MLSA to analyze real-world health data, this scenario — which is central to clinical practice — remains underexplored with a persistent gap in the systematic understanding of these approaches. In this context, this study represents one of the first in Brazil to apply and compare different MLSA algorithms to CRC data, contributing to the methodological advancement of the field and to the development of predictive tools with potential application in clinical practice and public health policies.

Compared to the group's previous research (Cardoso et al., 2023), which also used CRC data with classification ML models, this work represents an advance over its limitations. Although the previous algorithms showed good accuracy levels, they were constrained by the exclusion of a significant number of patients due to censoring (7%, 17.8%, and 26.9% for 1, 3, and 5-year survival prediction, respectively) — a common obstacle in long-term clinical studies or those based on real-world patient follow-up. The predictions from the two types of ML models indicated that classification ones underestimated survival probabilities, possibly due to patient exclusion, highlighting the advantage of MLSA in preserving complete cohort information.

In practice, the comparative analysis of MLSA algorithms reveals performance characteristics aligned with their theoretical foundations. RSF showed a consistent balance between predictive performance and computational cost but fell short of some boosting-based approaches. SSVM also demonstrated reasonable computational cost, although it yielded inferior results compared to other models. The survival analysis-adapted versions of XGB-Cox and LGBM did not achieve satisfactory metrics in this study, possibly due to limitations in their current implementations. Both XGB-AFT and GBSA achieved high prediction rates, with a small advantage for the first one.

The combined metrics allow a comprehensive evaluation of the models, considering both discriminatory capacity (via C-Index and C-Index IPCW) and reliability of survival estimates over time (via IBS and time-dependent AUC). However, since IBS and survival curves depend directly on the survival function, their application was restricted to RSF and GBSA, which incorporate it natively, and to SSVM, with an approximation via Cox model. Conversely, XGB-AFT, XGB-Cox, and LGBM algorithms do not have an appropriate estimation, limiting direct comparability with other models. Similarly, time-dependent AUC was calculated only for models that predict risks and was not estimated for XGB-AFT or LGBM.

During the hyperparameter optimization phase, computational cost proved to be a relevant challenge, especially considering the complexity of the models and the cohort size. A more cautious approach to this process was adopted, with the aim of increasing reliability in parameter selection. Although less costly alternatives could have been employed, this choice sought to prioritize consistency of predictive performance.

The SHAP and PI interpretability methods converged in identifying clinical staging as the most relevant feature, reinforcing methodological robustness. Their divergences, on the other hand, offer complementary perspectives and broaden the interpretation of results, as SHAP highlights local and individual effects, while PI emphasizes the global importance of variables. Whilst the analysis is restricted to the methods mentioned, it is relevant to mention SurvSHAP (Krzyziński, 2023), an extension of SHAP for survival analysis that represents a promising alternative for future investigations by explicitly considering censoring and providing interpretations of survival curves.

Recent MLSA literature has focused primarily on established models such as RSF and GBSA. Yang et al. (2023) applied these two algorithms to 2,157 colorectal cancer patients, using the C-Index and IBS as evaluation metrics, SHAP for interpretability, and Bayesian hyperparameter search with stratified 5-fold cross-validation. Andishgar et al. (2025) evaluated RSF and GBSA in 3,705 patients with traumatic brain injury, adopting C-Index, C-Index IPCW, IBS, and AUC metrics, with interpretability based on SHAP and Permutation Importance and optimization via 5-fold cross-validation. Fan et al. (2025) expanded the model set by including Survival SVM in 1,326 patients with epithelial ovarian cancer, using C-Index, IBS, and AUC metrics, SHAP-based interpretability, and hyperparameter tuning with 5-fold cross-validation. In this context, by applying a more comprehensive methodology to train and compare six MLSA algorithms on a registry of nearly 45,000 colorectal cancer patients, the present study demonstrates the models' performance and generalizability.

*Conclusion*

This study contributes to the systematization of the techniques discussed and reinforces the potential of MLSA algorithms not only in oncology but also in other contexts where predicting time to an event is relevant — such as industry, finance, and technology — thereby highlighting their cross-cutting applicability. Finally, we recommend that similar studies be conducted using survival neural networks, other cancer types, and with data from different regions, as well as on datasets that include richer clinical and sociodemographic information, in order to consolidate these methodologies as complementary tools for research and decision support.